%% file: paper.tex
\documentclass[10pt,conference]{IEEEtran}
\IEEEoverridecommandlockouts

\usepackage{cite}
\usepackage{amsmath,amssymb,amsfonts}
\usepackage{amsthm}
\usepackage{algpseudocode}
\usepackage{algorithm}
\usepackage{graphicx}
\usepackage{textcomp}
\usepackage{xcolor}
\usepackage{booktabs}
\usepackage{tabularx}
\usepackage{tcolorbox}
\usepackage{tikz}
\theoremstyle{definition}
\usepackage{subcaption}
\usepackage{xcolor}
\usepackage[hide links]{hyperref}
\definecolor{qmcolor}{RGB}{0,0,0}
\newcommand{\zqm}[1]{{#1}}
\newcommand{\method}{UML2Dep}

\newtheorem{definition}{Definition}
\def\BibTeX{{\rm B\kern-.05em{\sc i\kern-.025em b}\kern-.08em
    T\kern-.1667em\lower.7ex\hbox{E}\kern-.125emX}}
\begin{document}

\title{
Data Dependency-Aware Code Generation from Enhanced UML Sequence Diagrams
}

\author{
Wenxin Mao$^{1,\dagger,\ddagger}$, 
Zhitao Wang$^{1,\dagger}$,  
Long Wang$^{1,*}$, 
Sirong Chen$^{1,\ddagger}$, 
Cuiyun Gao$^{2,*}$, \\
Luyang Cao$^{1}$,  
Ziming Liu$^{1}$,  
Qiming Zhang$^{1}$,
Jun Zhou$^{1}$,
Zhi Jin$^{3}$ \\
$^1$ WeChat Pay, Tencent Inc, Shenzhen, China \\
$^2$ The Chinese University of Hong Kong, Hong Kong, China \\
$^3$ Wuhan University, Wuhan, China\\
mao23x@qq.com, \{zhitaowang, oliverlwang, sirongchen\}@tencent.com, cuiyungao@outlook.com, \\
\{lucascao, simonzmliu, archrzhang, anderszhou\}@tencent.com, zhijin@pku.edu.cn
\thanks{$^{\dagger}$These authors contributed equally to this work.}
\thanks{$^{\ddagger}$The work was done during an internship at Tencent.}
\thanks{$^{*}$These authors are the corresponding authors.}
}

\maketitle

\input{section/0_abstract}
\input{section/1_introduction}

\input{section/2_approach}
\input{section/3_experimental_setup}
\input{section/4_evaluation}
\input{section/5_system_demonstration}
\input{section/6_related_work}
\input{section/7_conclusion}
\input{section/8_ack}
\bibliographystyle{ieeetr}
\bibliography{ref}
\end{document}

%% file: section/0_abstract.tex
\begin{abstract}

Large language models (LLMs) excel at generating code from natural language (NL) descriptions. 
However, the plain textual descriptions are inherently ambiguous and often fail to capture complex requirements like intricate system behaviors, conditional logic, and architectural constraints; implicit data dependencies in service-oriented architectures are difficult to infer and handle correctly. 

To bridge this gap, we propose a novel step-by-step code generation framework named {\method} by leveraging unambiguous formal specifications of complex requirements. First, we introduce an enhanced Unified Modeling Language (UML) sequence diagram tailored for service-oriented architectures. This diagram extends traditional visual syntax by integrating decision tables and API specifications, explicitly formalizing structural relationships and business logic flows in service interactions to rigorously eliminate linguistic ambiguity. Second, recognizing the critical role of data flow, we introduce a dedicated data dependency inference (DDI) task. DDI systematically constructs an explicit data dependency graph prior to actual code synthesis. To ensure reliability, we formalize DDI as a constrained mathematical reasoning task through novel prompting strategies, aligning with LLMs’ excellent mathematical strengths. Additional static parsing and dependency pruning further reduce context complexity and cognitive load associated with intricate specifications, thereby enhancing reasoning accuracy and efficiency. 

Experimental results on our in-house industrial datasets demonstrate the effectiveness of the proposed framework. Specifically, our framework achieves strong performance, with 89.97\% recall, 95.06\% precision, and 92.33\% F1 score on the DDI task. Furthermore, the integration of {\method} into the code generation pipeline also improves practical deployment, increasing compilation pass rate by 8.83\% and unit test pass rate by 11.66\%.

\end{abstract}


%% file: section/1_introduction.tex
\section{Introduction}
Large language models (LLMs) have demonstrated remarkable capabilities in automated code synthesis from natural language (NL) descriptions, transforming how developers approach programming tasks~\cite{codex, hui2024qwen2, guo2024deepseek, JimenezYWYPPN24, codellama, gpt4o}. However, when scaling from simple demonstrations to production-grade software systems, a fundamental challenge persists: the inherent imprecision of natural language becomes increasingly problematic for capturing the detailed specifications required by complex applications. Natural language descriptions often lack the structural clarity needed to unambiguously define sophisticated system interactions, execution flows, and data relationships\cite{DBLP:journals/tse/FerrariGST22, DBLP:conf/birthday/GervasiFZS19}. 

Contemporary software engineering practice increasingly adopts Unified Modeling Language (UML) sequence diagrams as design blueprints\cite{shaikh2021more}. These diagrams provide formal visual syntax that explicitly captures control flows and service interactions with a obvious advancement over NL specifications, as demonstrated in model-driven engineering studies \cite{DBLP:conf/models/MussbacherABBKMWW14}. Recent work \cite{DBLP:conf/llm4code/AntalVF24,DBLP:conf/modelsward/SadikBO24} has systematically investigated LLM-based code generation from UML sequence diagrams, revealing persistent underperformance in industrial applications. The root cause lies in sequence diagrams' inherent limitation: while effectively modeling control logic, they lack explicit representation of data dependencies, i.e., the lifeblood of operational execution. This forces LLMs to simultaneously reconstruct control flow and infer implicit data dependencies during code generation, a dual cognitive load that obviously amplifies the difficulty and error likelihood of code generation. Particularly for large-scale sequence diagrams or exceptionally intricate logic, LLMs are highly susceptible to generating severe hallucinations when inferring data dependencies.

Addressing these practical challenges, we refocus on the core difficulty that hinders effective transformation from UML sequence diagrams to functional code: \zqm{the implicit expression of} data dependencies. This study proposes decoupling and preprocessing this challenge as an independent and critical task termed \textbf{Data Dependency Inference} (\textbf{DDI}) from UML sequence diagrams. \zqm{By} explicitly providing LLMs with \zqm{deterministic} control logic (derived from the sequence diagram) \zqm{and explicit data dependency information (obtained from upstream DDI), we substantially reduce LLMs' cognitive load associated with complex specifications, thereby ensuring the generation of high-quality code for industrial software.} Furthermore, it is noteworthy that DDI itself also holds obvious value during the software design phase, e.g., sequence diagram conformance checking and design validation \zqm{for software architects}.

To effectively resolve the critical DDI challenge, we propose a systematic solution framework \textbf{UML to Dependency-aware Code ({\method})} comprising:
\begin{itemize}
\item \textbf{Enhanced Sequence Diagram}:
Addressing the unique demands of complex software within service-oriented architecture (SOA), we propose an enhanced sequence diagram design specification. This specification not only encompasses standard sequence diagram elements but also integrates companion Decision Tables (explicitly defining business rules, validation logic, or workflow decisions) and granular API Specifications (detailing input parameters, outputs, data types, and constraints for each service interface). This rich, structured specification establishes a solid foundation for subsequent data dependency inference, enabling precise capture and expression of complex business rules and logic details.
\item \textbf{Mathematical Formalization Prompting}:
\zqm{To overcome the ambiguity of natural language prompts in complex DDI tasks requiring precise reasoning, we propose a novel prompting approach using formal mathematical descriptions. This method precisely defines inputs (e.g., sequence diagrams, decision tables, API specifications) and expected outputs (e.g., data dependency graphs) via mathematical expressions like functional dependencies and set operations. This structured and unambiguous approach directly leverages LLMs' inherent strength in mathematical reasoning, obviously enhancing their performance in generating complex data dependencies.}
\item \textbf{Reachability-Based Context Pruning}: 
Industrial sequence diagrams often generate extensive context. To alleviate LLM's cognitive load of DDI and enhance processing efficiency, we introduce a reachability-based context pruning technique. By parsing sequence diagrams into execution dependency graphs, we precisely identify reachable predecessor nodes for each target node in the execution flow, systematically removing logically unreachable context objects. This optimization strategy obviously reduces irrelevant noise presented to the LLM, enabling it to focus on constructing precise core data dependencies.
\end{itemize}

Our proposed framework {\method} obviously enhances sequence diagram quality and automated code generation for complex software systems. When evaluated on industrial datasets, the method achieved competitive performance in DDI task, notably attaining a recall of 89.97\%, an precision of 95.06\%, and an average F1 score of 92.33\%. 
Furthermore, incorporating DDI into the code generation pipeline obviously improved the compilation pass rate of generated code by 8.83\% and the unit test pass rate by 11.66\%. Based on these results, the framework has been successfully integrated into an industrial code generation pipeline, validating its effectiveness for practical deployment.

%% file: section/2_approach.tex
\section{PROPOSED FRAMEWORK}
\begin{figure*}
    \centering
    \includegraphics[width=0.9\textwidth]{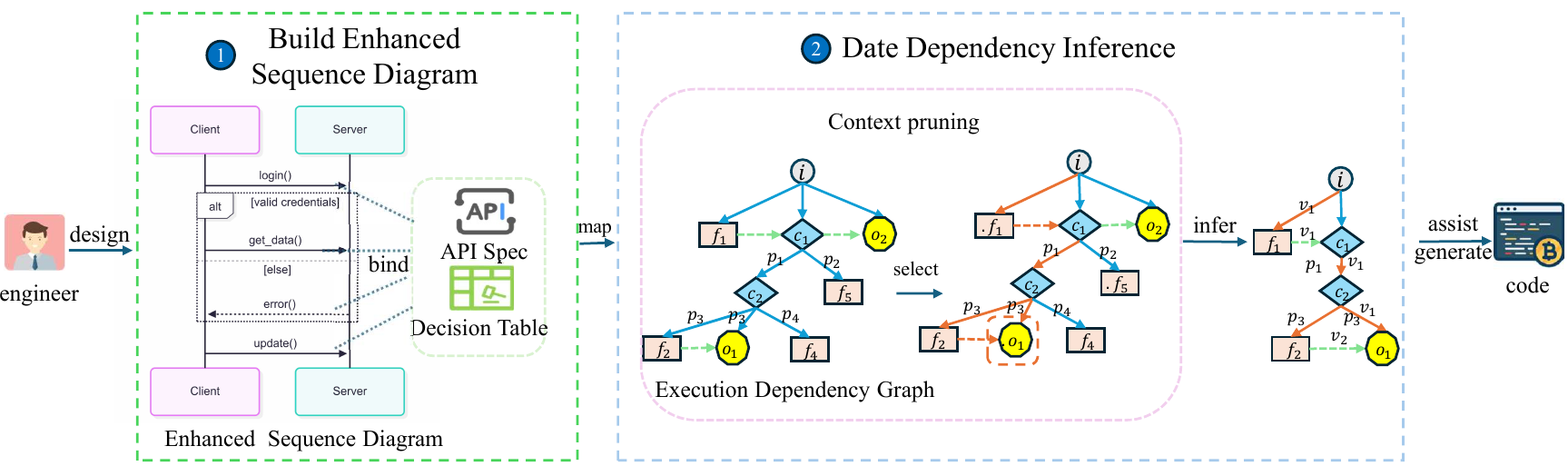}
    \caption{The overview of DDI Task and our framework}
    \label{fig:framework}
\end{figure*}
Figure \ref{fig:framework} demonstrates the overview of our framework {\method}. Engineers design the Enhanced Sequence Diagram based on specifications. After the context pruning process, we refine the available contextual information. This refined context, combined with mathematical formalization prompting, enables the LLMs to infer the data dependency graph. This graph is subsequently used to generate the final code.

\subsection{Enhanced Sequence Diagram}
Traditional methods for automatically generating code based on UML models~\cite{usman2008ujector,kluisritrakul2016generation} exhibit significant limitations when faced with complex software systems. The main reason is that classic sequence diagrams are unable to fully express complex business processes and data dependencies, resulting in generated code that only covers simple scenarios and fails to meet the requirements of industrial complex systems. In response to the unique characteristics of complex software, we propose an enhanced sequence diagram design specification to better support subsequent data dependency inference and high-quality code generation.

In this paper, we focus on software developed in the service-oriented architecture. In service-oriented architectures, UML sequence diagrams primarily model inter-service interactions through remote API calls, where each message typically represents a distributed service invocation rather than local method calls. Based on this characteristic, our specification retains the basic elements of UML sequence diagrams and introduces two key extensions:

\begin{itemize}
    \item \textbf{Decision Tables}: Used to clarify business logic constraints, boundary conditions, and exception handling paths. 
    Each \textbf{Decision Table} consists of a set of individual decision rules, and each rule is composed of two main components: \textbf{decision conditions} and \textbf{decision actions}.

    \begin{itemize}
        \item \textbf{Decision Condition}: This defines the logical predicate that must be satisfied for the corresponding execution action to be triggered.
        \item \textbf{Decision Action}: This specifies the operation to be performed when the condition is met (or unconditionally, if the condition is empty).
    \end{itemize}

    Both conditions and actions may reference or require specific data elements. Therefore, it is essential to analyze and infer the data dependencies associated with each decision rule---namely, to determine which data items are required to evaluate the conditions and to execute the actions.
    \item \textbf{Refined API Specifications}: These provide comprehensive API interface information encompassing: (1) detailed functional descriptions and application scenarios, (2) structured request/response formats with examples, and (3) precise data type definitions, business concepts, computation methods, and multiplicity constraints for each property. Such specifications ensure semantic precision for data fields through clear definitions, types, and constraints, while maintaining structural rigor for request/response contracts including mandatory/optional parameters and default values. The quality of these specifications directly impacts the accuracy of DDI.
\end{itemize}

In addition, for industrial scenarios, we propose the following design principles:

\begin{itemize}
    \item \textbf{Flexible Decision Table Binding}: In sequence diagrams, both messages and fragments can be bound to different categories of decision tables as needed by the business, enabling precise expression of conditional judgments, branch processing, and other complex logic.
    \item \textbf{Single Responsibility Constraint}: Each message is allowed to invoke only one API call, clarifying the functional boundary of each message, reducing coupling between messages, and improving the maintainability and readability of the sequence diagram.
    \item \textbf{One Use Case per Sequence Diagram}: Each sequence diagram should focus on expressing a single functional use case, i.e. the implementation of a single API. This ensures clarity and modularity in the design.
\end{itemize}

\subsection{Mathematical Formalization Prompting}\label{sec:formal-definition}

\subsubsection{Motivation}

Drawing inspiration from recent advances in the mathematical reasoning capabilities of Large Language Models (LLMs), we propose a mathematical formalization approach to enhance LLM performance on the Data Dependency Inference (DDI) task. The key insight is that DDI, fundamentally a graph-theoretic problem involving complex logical reasoning about data flow relationships, \zqm{can benefit significantly from rigorous mathematical abstractions.
This mathematical foundation provides a structured framework for systematic reasoning about data dependencies, ensuring completeness and correctness of the inference process.}
Furthermore, the mathematical formulation allows us to establish clear constraints and validation criteria for dependency relationships, enabling more reliable evaluation and verification of LLM-generated results.

\begin{figure}[ht]
    \centering
    \begin{tcolorbox}[colback=gray!10,colframe=gray!50,title=DDI Mathematical Formalization Prompt Template, coltitle=black]
    \scriptsize
    \textbf{\# Formal Problem Specification}
    \\
    \texttt{Given a UML sequence diagram, construct a data dependency graph}
    \\
    \texttt{$\mathcal{G}_{\text{DD}} = (\mathcal{V}, \mathcal{E}_{\text{DD}}, \mathcal{D})$ where:}
    \\
    \texttt{- Nodes: [Node Definition]}
    \\
    \texttt{- Edges: [Edge Definition]}
    \\
    \texttt{- Data consumption categories: [Consumption definition]}
    \\
    \texttt{- Data production categories: [Production definition]}
    \\
    
    \textbf{\# Contextual Information}
    \\
    \texttt{**Sequence Diagram Context:**}
    \\
    \texttt{[Usecase description and background information]}
    \\
    
    \texttt{**Reachable Nodes $\mathcal{P}(t)$:**}
    \\
    \texttt{For each node $s \in \mathcal{P}(t)$:}
    \\
    \texttt{- Type: [Input/Function/Control]}
    \\
    \texttt{- API: [API specs]}
    \\
    \texttt{- Decision Tables: [Conditions/Actions]}
    \\
    
    \texttt{**Target Node $t$:**}
    \\
    \texttt{- Type: [Function/Control/Output]}
    \\
    \texttt{- API: [API specs]}
    \\
    \texttt{- Decision Tables: [Conditions/Actions]}
    \\
    
    \textbf{\# Inference Constraints}
    \\
    \texttt{[Completeness requirements, dependency path validity, consistency]}
    \\
    
    \textbf{\# Output Format}
    \\
    \texttt{Provide dependency edges as follows: }
    \\
    \texttt{[JSON schema]}
    \end{tcolorbox}
    \caption{Structured prompt template for DDI mathematical formalization}
    \label{fig:prompt-template}
    \end{figure}

\subsubsection{Prompt Structure}

The mathematical formulation is integrated into a structured prompt structure designed to elicit optimal reasoning performance from LLMs on the DDI task. \zqm{The prompt template follows a systematic four-component structure as illustrated in Fig.~\ref{fig:prompt-template}.}

Each component serves a specific purpose in guiding LLM reasoning, \zqm{and the detailed mathematical definition can be found in the following section.}
\begin{itemize}
    \item \textbf{Formal Problem Specification} \zqm{establishes} the mathematical foundation and constrains the solution space through rigorous definitions
    \item \textbf{Contextual Information} \zqm{provides} structured domain knowledge organized into three categories: sequence diagram context, candidate dependent nodes $\mathcal{P}(t) = \{s \in \mathcal{V} \setminus \{t\} : \text{reachable}(s, t)\}$, and target node specifications
    \item \textbf{Inference Constraints} \zqm{ensures} data completeness, dependency path validity, and consistency with the reachability relation $\text{reachable}(s, t)$
    \item \textbf{Output Format Specification} \zqm{guarantees} machine-readable results that conform to the mathematical definition
\end{itemize}

\subsubsection{Mathematical Definition of DDI Task}


\begin{definition}[Data Dependency Inference Problem]
Given a UML sequence diagram $\mathcal{SD} = (M, F)$, where $M$ denotes the set of messages and $F$ denotes the set of interaction fragments, the DDI task aims to construct a directed data dependency graph $\mathcal{G}_{\text{DD}} = (\mathcal{V}, \mathcal{E}_{\text{DD}}, \mathcal{D})$ that captures all data flow relationships within the system.
\end{definition}

\begin{definition}[Data Dependency Graph]
A data dependency graph $\mathcal{G}_{\text{DD}} = (\mathcal{V}, \mathcal{E}_{\text{DD}}, \mathcal{D})$ is a directed graph where:
\begin{itemize}
    \item $\mathcal{V}$ is the vertex set representing computational and control \zqm{nodes} in the sequence diagram.
    \item $\mathcal{E}_{\text{DD}} \subseteq \mathcal{V} \times \mathcal{D} \times \mathcal{V}$ is the edge set representing data flow relationships, where each edge $(s, d, t) \in \mathcal{E}_{\text{DD}}$ indicates that node $s$ produces data entity $d$ that is consumed by node $t$.
    \item $\mathcal{D}$ is the domain of all possible data entities that can flow between nodes.
\end{itemize}
\end{definition}

\begin{definition}[Data Dependency Node]
The node set $\mathcal{V}$ is formally partitioned into four disjoint subsets:
$$\mathcal{V} = \mathcal{F} \cup \mathcal{C} \cup \mathcal{I} \cup \mathcal{O}$$
where $\mathcal{F} \cap \mathcal{C} \cap \mathcal{I} \cap \mathcal{O} = \emptyset$, and each subset is defined as follows:

\begin{itemize}
    \item $\mathcal{F}$ (\textbf{Function Node Set}): Each node $f \in \mathcal{F}$ corresponds to a standard \emph{Message} in the UML sequence diagram, excluding \emph{Return Messages}. These messages represent function or method invocations between objects, responsible for executing computational logic or data processing operations (e.g., remote API calls, decision-based operations), and generating output data for subsequent processing nodes.
    
    \item $\mathcal{C}$ (\textbf{Control Node Set}): Each node $c \in \mathcal{C}$ corresponds to an \emph{Interaction Fragment} in the UML sequence diagram, including \texttt{opt}, \texttt{alt}, \texttt{loop}, and \texttt{break} constructs. These fragments represent control flow decision points (e.g., conditional branches, iterative structures, or early termination conditions) that evaluate boolean or selection predicates based on input data to determine subsequent execution paths. The outgoing edges of control nodes typically represent alternative execution branches, reflecting divergent control flows.
    
    \item $\mathcal{I}$ (\textbf{Input Node Set}): Typically $|\mathcal{I}| = 1$ and $\forall i \in \mathcal{I}: \text{in-degree}(i) = 0$. This set contains a single virtual node. \zqm{The node has all necessary external input data for the entire functional use case represented by the sequence diagram.}
    
    \item $\mathcal{O}$ (\textbf{Output Node Set}): Each node $o \in \mathcal{O}$ corresponds to a \emph{Return Message} in the UML sequence diagram, \zqm{responsible for collecting the final output data of the functional use case.} Formally, $\forall o \in \mathcal{O}: \text{out-degree}(o) = 0$, as outputs are directed to external consumers and have no internal dependencies.
\end{itemize}
\end{definition}

\begin{definition}[Data Dependency Edge]
The edge set $\mathcal{E}_{\text{DD}}$ represents data dependency relationships between reachable nodes. Each edge is formally defined as a triple $(s, d, t) \in \mathcal{E}_{\text{DD}}$, where:
\begin{itemize}
    \item $s \in \mathcal{V}$ is the \emph{source node} (data producer)
    \item $d$ is the \emph{\zqm{data}} transmitted along the edge
    \item $t \in \mathcal{V}$ is the \emph{target node} (data consumer)
\end{itemize}
\end{definition}

The edge set $\mathcal{E}_{\text{DD}}$ satisfies the following formal constraints:
\begin{align*}
\forall (s, d, t) \in \mathcal{E}_{\text{DD}}: \text{reachable}(s, t), \\
s \in (\mathcal{F} \cup \mathcal{C} \cup \mathcal{I}), t \in (\mathcal{F} \cup \mathcal{C} \cup \mathcal{O}) 
\end{align*}

Finally, we formalize the data consumption and production mechanisms that drive dependency inference:
\begin{definition}[Data Consumption and Production Categories]
For systematic dependency analysis, we \zqm{categorize subjects within each node as either data producers or data consumers. This enables the LLM to analyze the data dependency step by step:
$$\mathcal{D}_{\text{produce}}(s) = \mathcal{D}_{\text{API-Resp}}(s) \cup \mathcal{D}_{\text{DecisionTable-Out}}(s)$$
$$\mathcal{D}_{\text{consume}}(t) = \mathcal{D}_{\text{API-Req}}(t) \cup \mathcal{D}_{\text{DecisionTable-In}}(t)$$
where \textit{API-Resp} and \textit{API-Req} denote the response and request data of the API, while \textit{DecisionTable-Out} and \textit{DecisionTable-In} denote the input and output data for decision tables.}
\end{definition}

\subsection{Reachability-Based Context Pruning}

\zqm{A straightforward approach to organizing the target node's DDI context is to consider all preceding nodes in the sequence diagram $\mathcal{SD}$ as potential data suppliers.} However, this approach suffers from critical limitations in practical deployment scenarios:

\begin{itemize}
    \item \textbf{Scalability Issues}: Industrial business logic often \zqm{shows significant complexity} with extensive node sets $|\mathcal{V}|$, causing context length to exceed LLM processing capabilities.
    \item \textbf{Cognitive Overload}: Inclusion of irrelevant contextual information increases computational burden and may introduce spurious correlations that degrade inference accuracy.
    \item \textbf{Noise Amplification}: Redundant context objects can mislead the model with incorrect cues, compromising the precision of dependency edge construction.
\end{itemize}

To address these challenges while preserving the theoretical soundness of our DDI formulation, we propose a \emph{reachability-based context pruning strategy}. It leverages the execution reachability relation $\text{reachable}(s, t)$ defined \zqm{in the following.

\begin{definition}[Execution Reachability]
A node $s \in \mathcal{V}$ is \emph{execution-reachable} from node $t \in \mathcal{V}$, denoted as $\text{reachable}(s, t)$, if there exists at least one feasible execution path $\pi$ such that execution can transition from node $s$ to node $t$:
$$\text{reachable}(s, t) \iff \exists \pi \in \Pi: s \xrightarrow{\pi} t$$
where $\Pi$ denotes the set of all possible execution paths. Note that input nodes are reachable from all nodes: $\forall t \in \mathcal{V} \setminus \mathcal{I}, \forall i \in \mathcal{I}: \text{reachable}(i, t) = \text{True}$.
\end{definition}
}

\begin{definition}[Context Pruning Problem]
Given a target node $t \in \mathcal{V}$ in the data dependency graph $\mathcal{G}_{\text{DD}} = (\mathcal{V}, \mathcal{E}_{\text{DD}}, \mathcal{D})$, identify the minimal predecessor node set $\mathcal{P}(t) \subseteq \mathcal{V}$ such that:
\begin{align*}
\mathcal{P}(t) &= \{s \in \mathcal{V} \setminus \{t\} : \text{reachable}(s, t)\}, \\
&\forall (s, d, t) \in \mathcal{E}_{\text{DD}}: s \in \mathcal{P}(t) \label{eq:completeness}
\end{align*}
where the predecessor set $\mathcal{P}(t)$ contains all nodes that can reach the target node $t$ through execution flow paths, excluding the target node itself. Since the input node set $\mathcal{I}$ is reachable from all other nodes, it is naturally included in this set. Above constraint ensures that all potential data sources for target node $t$ are contained within the predecessor set, guaranteeing completeness of the pruned context. The effective search space for LLM reasoning is reduced from $|\mathcal{V}|$ to $|\mathcal{P}(t)|$, with token consumption decreasing proportionally to $\frac{|\mathcal{P}(t)|}{|\mathcal{V}|}$. Additionally, Systematic exclusion of irrelevant context enables LLM attention mechanisms to concentrate on essential dependency relationships, reducing hallucination probability and noise interference.
\end{definition} 

\subsubsection{EDG}

To operationalize the execution reachability relation, we introduce the \emph{Execution Dependency Graph} (EDG), a specialized graph structure that captures both hierarchical nesting and temporal execution semantics inherent in UML sequence diagrams.

\begin{definition}[EDG]
Given a sequence diagram $\mathcal{SD} = (M, F)$, the corresponding EDG is defined as a directed graph $\mathcal{G}_{\text{ED}} = (\mathcal{V}, \mathcal{E}_{\text{ED}})$, where:
\begin{itemize}
    \item $\mathcal{V} = \mathcal{F} \cup \mathcal{C} \cup \mathcal{I} \cup \mathcal{O}$ is the same node set as defined in the Data Dependency Graph, representing the four types of computational and control entities. The input node set $\mathcal{I}$ serves as the root of the EDG, ensuring all nodes in the graph are reachable from $\mathcal{I}$.
    \item $\mathcal{E}_{\text{ED}} = \mathcal{E}_H \cup \mathcal{E}_S$ where:
    \begin{itemize}
        \item $\mathcal{E}_H \subseteq \mathcal{V} \times \mathcal{V}$: hierarchical containment edges (parent-child relationships)
        \item $\mathcal{E}_S \subseteq \mathcal{V} \times \mathcal{V}$: sequential execution edges (temporal precedence relationships)
    \end{itemize}
\end{itemize}
\end{definition}

\begin{figure*}
    \centering
    \includegraphics[width=0.8\textwidth]{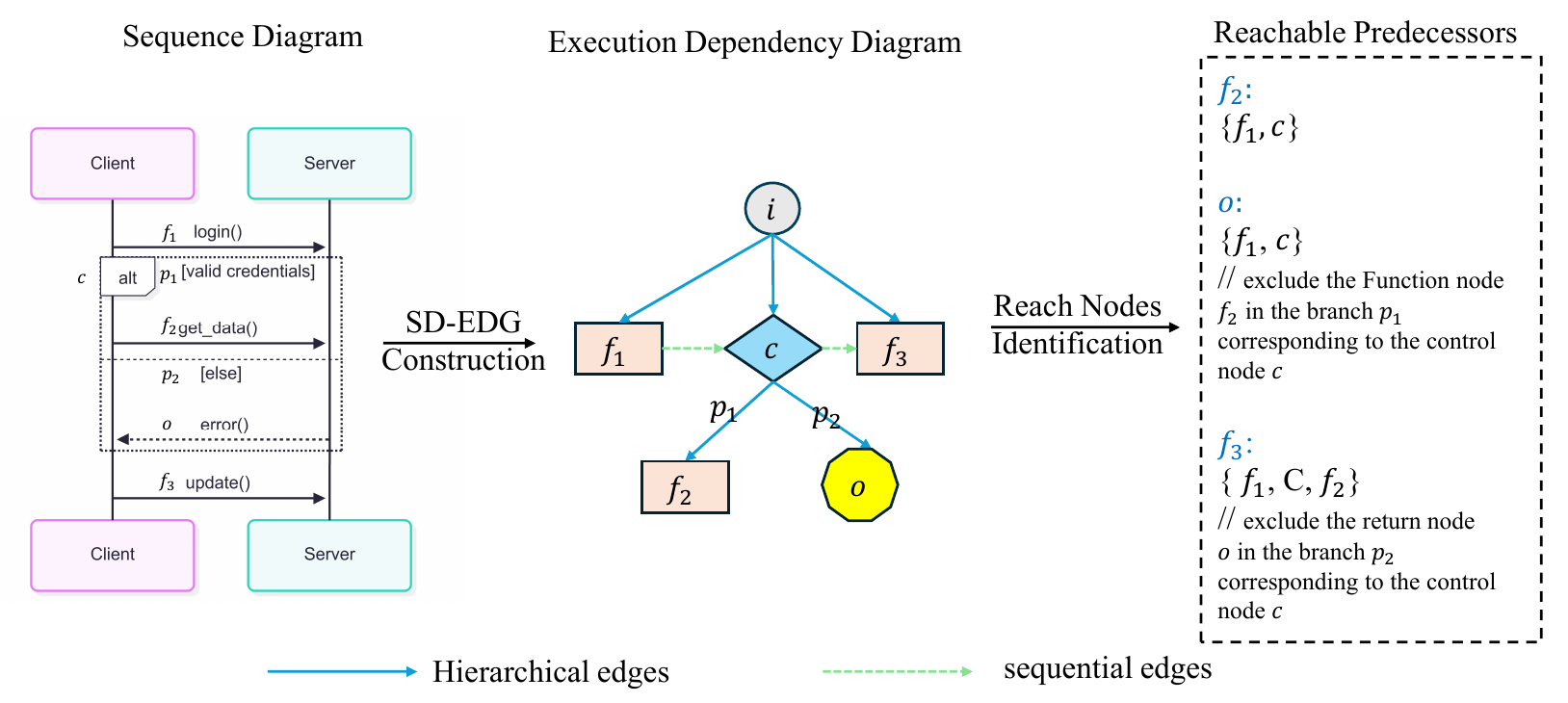}
    \caption{EDG Construction and Reachable Predecessors Identification on EDG}
    \label{fig:sd-to-edg}
\end{figure*}

The EDG preserves the structural semantics of sequence diagrams through its dual-edge architecture: hierarchical edges $\mathcal{E}_H$ capture the nesting relationships between interaction fragments and their contained elements, while sequential edges $\mathcal{E}_S$ encode the temporal execution order within each hierarchical scope.
Figure~\ref{fig:sd-to-edg} demonstrates the systematic transformation from sequence diagram visual representation to the corresponding EDG structure.

\subsubsection{EDG Construction}

The construction of EDG follows a systematic dual-phase methodology that decomposes the transformation process into hierarchical structure extraction and sequential relationship inference. The input node set $\mathcal{I}$ is established as the root of the EDG, serving as the starting point from which all other nodes become reachable through execution paths.

Hierarchical Relationship Construction: We establish the hierarchical edge set $\mathcal{E}_H$ through spatial containment analysis. For any container fragment $p \in \mathcal{C}$, we define the spatial containment relation function $\text{contains}: \mathcal{C} \times (\mathcal{F} \cup \mathcal{C}) \rightarrow \{0,1\}$, where $\text{contains}(p, e) = 1$ if and only if element $e$ is spatially enclosed within the visual boundaries of fragment $p$.

Sequential Relationship Construction: Following the establishment of hierarchical structure, we construct the sequential edge set $\mathcal{E}_S$ through recursive depth-first traversal. For any node $v \in \mathcal{V}$, we define its direct child set as $\text{children}(v) = \{u \in \mathcal{V} : (v, u) \in \mathcal{E}_H\}$. For a temporally ordered child sequence within each hierarchical scope, sequential edges are constructed between consecutive elements that are not mutually exclusive. The key constraint is that alternative branches within interaction fragments (e.g., \texttt{alt}) represent mutually exclusive execution paths rather than sequential dependencies. 

Recursive Construction Strategy: The construction process employs a top-down recursive strategy that ensures proper establishment of sequential relationships at each hierarchical level. For each container fragment $p \in \mathcal{C}$, after completing the sequential edge construction among its direct children, the same construction process is recursively applied to all its child container fragments. This recursive methodology guarantees comprehensive capture of sequential relationships across all levels of nested structures.

\begin{algorithm}[H]
\caption{Reachable Predecessor Identification}
\begin{algorithmic}[1]
\scriptsize
\Require Target node $t$, EDG $\mathcal{G}_{\text{ED}} = (\mathcal{V}, \mathcal{E}_{\text{ED}})$
\Ensure Reachable predecessor set $\mathcal{P}(t)$
\State $\mathcal{R} \leftarrow \emptyset$, $\mathcal{U} \leftarrow \emptyset$
\State $\texttt{backward\_traversal}(t, \mathcal{R}, \mathcal{U})$
\State $\mathcal{R} \leftarrow \texttt{filter\_return\_branches}(\mathcal{R})$
\State $\mathcal{P}(t) \leftarrow \mathcal{R} \setminus \{t\}$
\Return $\mathcal{P}(t)$

\Procedure{backward\_traversal}{node, $\mathcal{R}$, $\mathcal{U}$}
    \If{$node \in \mathcal{U}$} \Return \EndIf
    \State $\mathcal{U} \leftarrow \mathcal{U} \cup \{node\}, \mathcal{R} \leftarrow \mathcal{R} \cup \{node\}$
    \For{each $(p, node) \in \mathcal{E}_H$}
        \State $\texttt{backward\_traversal}(p, \mathcal{R}, \mathcal{U})$
    \EndFor
    \For{each $(s, node) \in \mathcal{E}_S$}
        \State $\texttt{backward\_traversal}(s, \mathcal{R}, \mathcal{U})$
        \State $\texttt{explore\_subtree}(s, \mathcal{R}, \mathcal{U})$
    \EndFor
\EndProcedure

\Procedure{explore\_subtree}{node, $\mathcal{R}$, $\mathcal{U}$}
    \For{each $(node, c) \in \mathcal{E}_H$}
        \If{$c \notin \mathcal{U}$ \textbf{and} $\neg\texttt{is\_return\_branch}(c)$}
            \State $\texttt{backward\_traversal}(c, \mathcal{R}, \mathcal{U})$
        \EndIf
    \EndFor
\EndProcedure
\end{algorithmic}
\end{algorithm}

\subsubsection{Reachable Predecessor Identification}
Building upon the EDG structure and the execution reachability definition, the predecessor identification algorithm performs comprehensive backward traversal of the EDG, systematically exploring both hierarchical and sequential dependencies while filtering non-contributing return branches.

The algorithm starts by initializing reachable node set $\mathcal{R}$ and visited node set $\mathcal{U}$, then performs backward exploration from target node $t$. During traversal, it explores hierarchical parent relationships through edges in $\mathcal{E}_H$ to capture containment dependencies, and examines sequential predecessor relationships through edges in $\mathcal{E}_S$. For each predecessor node, the algorithm recursively applies backward traversal and explores all children through subtree analysis to capture nested elements. Since the input node set $\mathcal{I}$ serves as the root of the EDG, the backward traversal will ultimately reach $\mathcal{I}$ for any reachable target node, ensuring completeness of the predecessor set. Finally, it applies filtering mechanisms to exclude return branches, remove the target node itself.

%% file: section/3_experimental_setup.tex
\section{EXPERIMENTAL SETUP}

\input{tables/dependency_detail}

\subsection{Datasets}
We collect 11 design sequence diagrams from internal business data of WeChat Pay and extract their dependency relations to construct our dataset. The dataset contains a total of 224 dependency relations, including 157 API dependencies, 60 decision condition dependencies, and 7 decision action dependencies. Each dependency relation is reviewed and approved by three employees. Our dependency inference framework has been deployed on the internal business platform to assist designers in verifying designs and to enhance the effectiveness of code generation. The details of the dependency relations for each use case in the dataset are presented in Table~\ref{tab:dependency_count}.

\subsection{Evaluation Metrics} 
To evaluate the correctness of the inferred dependency relations, we adopt three standard metrics: Precision, Recall, and F1 score. 

\textbf{Precision} measures the proportion of correctly predicted dependency edges among all edges generated by the model, reflecting the accuracy of the predictions. 
\textbf{Recall} quantifies the proportion of ground-truth dependency edges that are successfully identified by the model, indicating the completeness of the inference. 
The \textbf{F1 score} is the harmonic mean of Precision and Recall, providing a balanced assessment of both accuracy and completeness.

In our practical business scenario, it is crucial to ensure that all true dependency edges are identified (i.e., high Recall), as missing dependencies may lead to critical design or implementation errors. Once completeness is ensured, we further focus on improving the correctness of the predicted edges (i.e., high Precision).
\subsection{Model and hyper-parameter}
Due to confidentiality requirements regarding internal business data, we select the open-source models \texttt{deepseek-r1-0528}\cite{deepseekr1} and \texttt{deepseek-v3-0324}\cite{deepseekv3} for our experiments and deploy them on our own infrastructure. For all experiments, we set the temperature parameter to 0.1 to ensure more deterministic and stable outputs.

%% file: tables/dependency_detail.tex
\begin{table}[htbp]
\footnotesize
\centering
\caption{Statistics on the number of nodes and dependency edges for each use case}
\label{tab:dependency_count}
\resizebox{0.45\textwidth}{!}{\begin{tabular}{l|ccc|ccc}
\toprule
\textbf{UseCase} 
& \multicolumn{3}{c|}{\textbf{Dependency Edges}} 
& \multicolumn{3}{c}{\textbf{Nodes}} \\
\cmidrule(lr){2-4} \cmidrule(lr){5-7}
& \textbf{API} & \textbf{Condition} & \textbf{Action} 
& $|\mathcal{F}|$ & $|\mathcal{C}|$ & $|\mathcal{O}|$ \\
\midrule
ClearFlag                 & 1  & 0 & 0 & 1 & 0 & 1 \\
SetFlag                   & 1  & 0 & 0 & 1 & 0 & 1 \\
QueryParentAccounts & 2  & 0 & 0 & 1 & 0 & 1 \\
BindCard    & 6  & 0 & 0 & 1 & 0 & 1 \\
SetPassiveLimit & 9  & 5 & 0 & 5 & 2 & 1 \\
SetActiveLimit  & 11 & 5 & 0 & 5 & 2 & 1 \\
VerifyUserFace & 26 & 0 & 0 & 2 & 0 & 1 \\
SetAccountDailyQuota & 23 & 4 & 1 & 6 & 1 & 1 \\
SetPayKey                 & 18 & 5 & 3 & 7 & 2 & 3 \\
QueryPMAccount       & 27 & 2 & 0 & 3 & 1 & 1 \\
OpenPSAccount       & 33 & 39 & 3 & 18 & 4 & 3 \\
\midrule
Overall                       & 157 & 60 & 7 & 50 & 12 & 15 \\
\bottomrule
\end{tabular}
}
\end{table}

%% file: section/4_evaluation.tex
\section{EVALUATION}

\input{tables/perfermance_on_DDI}

\input{tables/comparison_of_MathPrompt}

\input{tables/comparison_of_context_purning}

In this section, we present the research questions that guide our evaluation:

\begin{itemize}
    \item \textbf{RQ1:} How effective is {\method} for the Data Dependency Inference (DDI) task?
    \item \textbf{RQ2:} How effective are the individual components of {\method}?
    \item \textbf{RQ3:} How useful is {\method} in real-world scenarios for assisting code generation?
\end{itemize}

\subsection{RQ1: How effective is {\method} for the Data Dependency Inference (DDI) task?}

To evaluate the effectiveness of our framework {\method} on the DDI task, we analyze the results of both \texttt{deepseek-r1} and \texttt{deepseek-v3} models across all use cases in our dataset. Table~\ref{tab:dependency_count} presents the number of dependencies for each use case, which reflects the complexity of the corresponding sequence diagrams. Table~\ref{tab:performance_comparison} report the detailed performance metrics (Precision, Recall, and F1) for each use case and dependency type.

\paragraph{Overall Effectiveness}
Both models achieve high overall performance on the DDI task. For most use cases, the F1 scores exceed 80\%, and in many cases, they reach above 90\%. This demonstrates that our framework can accurately infer data dependencies from industrial-scale sequence diagrams, even when the diagrams are complex and contain a large number of dependencies.

\paragraph{Impact of Use Case Complexity}
A detailed analysis of the results underscores the significant impact of use case complexity—quantified by both the number of dependencies and nodes—on model performance. For simple use cases such as \textit{ClearFlag}, \textit{SetFlag}, \textit{QueryParentAccounts}, and \textit{BindCard}, which contain few dependencies and nodes, both \texttt{deepseek-v3} and \texttt{deepseek-r1} achieve perfect scores across all evaluation metrics. The limited structural complexity in these scenarios enables both models to reliably infer all data dependencies.

For use cases of moderate complexity, such as \textit{SetPassiveLimit} and \textit{SetActiveLimit}, we observe that \texttt{deepseek-v3} frequently outperforms \texttt{deepseek-r1}. These cases typically involve a moderate number of dependencies and several branching paths. \texttt{deepseek-v3} efficiently infers the correct dependencies after capturing the overall logic, whereas \texttt{deepseek-r1} often overanalyzes possible branches, leading to inconsistent conclusions and reduced performance.

A notable exception is \textit{QueryPMAccount}, which, despite containing relatively few nodes, exhibits high complexity due to the large number of dependencies associated with a single API request node. In this case, \texttt{deepseek-r1} reasons through each parameter individually, resulting in a prolonged and sometimes inconsistent inference process that ultimately degrades its performance. In contrast, as a non-reasoning model, \texttt{deepseek-v3} efficiently infers all dependencies in aggregate, leading to superior efficiency and accuracy in this scenario.

For highly complex use cases such as \textit{SetAccountDailyQuota}, \textit{SetPayKey}, and \textit{OpenPSAccount}—characterized by many dependencies, numerous nodes, and multiple branching execution paths—\texttt{deepseek-v3} struggles to maintain high performance. In these challenging scenarios, \texttt{deepseek-r1} consistently achieves higher recall and F1 scores, leveraging its advanced reasoning capabilities to accurately capture intricate and implicit data dependencies across complex control flows.

\begin{tcolorbox}
\textbf{Answer to RQ1:} 
{\method} achieves an average recall of 89.97\% in data dependency inference across all use cases. For simple and moderately complex sequence diagrams, \texttt{deepseek-v3} performs comparably to \texttt{deepseek-r1} while providing faster inference, making it preferable in most practical scenarios. For complex diagrams with many dependencies, \texttt{deepseek-r1} offers superior reasoning and higher accuracy, and is recommended for critical or large-scale design tasks.
\end{tcolorbox}

\subsection{RQ2: How effective are the individual components of {\method}?}

To assess the effectiveness of each component in our framework, we compare three settings on the DDI task: (1) using mathematical formalization prompting with context pruning (our default method), (2) using natural language description, and (3) using mathematical formalization prompting without context pruning. The results are shown in Table~\ref{tab:performance_comparison}, Table~\ref{tab:math_prompt_comparison}, and Figure~\ref{fig:rbcp}, respectively. We focus our analysis on the six use cases with redundant nodes and complex control flows, as context pruning is only applicable in these scenarios.

\paragraph{Impact of Mathematical Formalization Prompting}
Across all complex use cases, mathematical formalization prompting consistently outperform natural language descriptions in terms of F1 score, which is our primary metric to ensure that no dependencies are missed. For example, in the \textit{SetPayKey} use case, the F1 score with mathematical formalization prompting is 82.18\%, compared to 72.51\% with natural language. The advantage is even more pronounced in the \textit{SetAccountDailyQuota} use case, where the F1 score drops from 80.32\% (formalism) to 74.84\% (natural language). This demonstrates that precise, structured input enables the model to better capture complex data dependencies, especially when both API and Condition dependencies are present.

\paragraph{Effectiveness of Reachability-Based Context Pruning}
As shown in the Figure~\ref{fig:rbcp}, Reachability-Based Context Pruning (RBCP) leads to overall improvements in precision, recall, and F1 score across most use cases. CP generally enhances recall, particularly in complex cases such as \textit{SetPayKey} and \textit{SetPassiveLimit}, indicating better coverage of relevant dependencies. Precision is also maintained or slightly improved, showing that pruning does not introduce more false positives. Overall, CP contributes to consistently higher or comparable F1 scores, demonstrating its effectiveness in filtering irrelevant context.

\begin{tcolorbox}
\textbf{Answer to RQ2:} 
Each component of {\method} contributes to enhancing data dependency inference (DDI) performance. mathematical formalization prompting provides more precise and structured input, effectively reducing the solution space and improving inference, particularly for complex dependencies. In scenarios with numerous dependencies and redundant control paths, context pruning mitigates the impact of irrelevant information, thereby improving inference accuracy and efficiency. 

\end{tcolorbox}

\subsection{How useful is {\method} in real-world scenarios for 
assisting code generation?}
\input{tables/codegen}

To assess the real-world effectiveness of {\method}, we conducted experiments on six representative use cases drawn from WeChat Pay’s internal business system. These cases involve enterprise-level, customized code frameworks, under which general-purpose large language models often struggle to generate syntactically correct and functionally complete code. Our method introduces data dependency information into the generation process. This structured information enables the construction of a Data Flow Graph (DFG) that guides the generation of function signatures and call relationships, helping the model align with expected control and data flows. As a result, the generated code is more likely to compile and pass unit tests.

Table~\ref{tab:compile_unittest_comparison_diff_reordered} presents a comparative evaluation under three quality metrics. 
\begin{itemize}
    \item The \textbf{Compilation Pass Rate} indicates whether the generated code can be compiled successfully, reflecting syntactic and structural correctness.
    \item The \textbf{Full Unit Test Pass Rate} reflects the percentage of code that passes all unit tests, representing complete functional accuracy. 
    \item The \textbf{Unit Test Pass Rate} reflects the average test coverage passed, indicating partial correctness.
\end{itemize}

From the data, we observe consistent improvements across all three metrics when data dependency information is used. Compilation pass rate improves from 85.50\% to 94.33\%, a gain of 8.83 percentage points. Full unit test pass rate increases by 5 percentage points, while the unit test pass rate improves even more significantly, from 81.17\% to 92.83\%—an increase of 11.66 percentage points. These gains are not just statistical; they indicate that the model is generating more structurally sound code that better reflects business intent.

The benefit of data dependencies becomes especially apparent in more complex use cases. For instance, in \texttt{QueryPMAccount}, compilation pass rate improves by 29 percentage points, and unit test pass rate increases by 36 points. This suggests that, in scenarios where logic is deeply intertwined with upstream/downstream data flows, the availability of structured dependency context is essential for generating coherent and executable code. In \texttt{SetAccountDailyQuota}, the full unit test pass rate jumps from 0\% to 27\%, showing that the model struggles to assemble a complete functional unit without data flow guidance.

While there are rare exceptions, such as \texttt{QueryParentAccounts} showing a slight drop in full unit test pass rate, the overall trend is robust. These findings suggest that data dependency information plays a foundational role in bridging the gap between local token-level generation and the global structural correctness required by real-world applications.

\begin{tcolorbox}
\textbf{Answer to RQ3:} 
Providing data dependency information enables models to construct accurate data flow graphs and function signatures, significantly improving compilation success and test pass rates in complex real-world code generation scenarios.
\end{tcolorbox}

%% file: tables/perfermance_on_DDI.tex
\begin{table*}[htbp]
\centering
\vspace{-1em}
\caption{Comparison of DDI task results based on deepseek-r1 and deepseek-v3. The results are represented as data1/data2, where data1 and data2 refer to deepSeek-r1 and deepSeek-v3 results, respectively. The higher value is bolded.}
\label{tab:performance_comparison}
\resizebox{\textwidth}{!}{
\begin{tabular}{@{}l|ccc|ccc|ccc|ccc@{}}
\toprule
\textbf{UseCase} 
& \multicolumn{3}{c|}{Overall} 
& \multicolumn{3}{c|}{API} 
& \multicolumn{3}{c|}{Condition} 
& \multicolumn{3}{c}{Action} \\
\cmidrule(lr){2-4} \cmidrule(lr){5-7} \cmidrule(lr){8-10} \cmidrule(lr){11-13}
& Precision & Recall & F1 
& Precision & Recall & F1 
& Precision & Recall & F1 
& Precision & Recall & F1 \\
\midrule
ClearFlag 
& \textbf{100.00}/100.00 & \textbf{100.00}/100.00 & \textbf{100.00}/100.00 
& \textbf{100.00}/100.00 & \textbf{100.00}/100.00 & \textbf{100.00}/100.00 
& - & - & - & - & - & - \\
SetFlag 
& \textbf{100.00}/100.00 & \textbf{100.00}/100.00 & \textbf{100.00}/100.00 
& \textbf{100.00}/100.00 & \textbf{100.00}/100.00 & \textbf{100.00}/100.00 
& - & - & - & - & - & - \\
QueryParentAccounts 
& \textbf{100.00}/83.64 & \textbf{100.00}/100.00 & \textbf{100.00}/86.15 
& \textbf{100.00}/83.64 & \textbf{100.00}/100.00 & \textbf{100.00}/86.15 
& - & - & - & - & - & - \\
BindCard 
& \textbf{100.00}/100.00 & \textbf{100.00}/100.00 & \textbf{100.00}/100.00 
& \textbf{100.00}/100.00 & \textbf{100.00}/100.00 & \textbf{100.00}/100.00 
& - & - & - & - & - & - \\
SetPassiveLimit 
& \textbf{93.31}/90.28 & 78.57/\textbf{87.14} & 85.23/\textbf{88.31} 
& \textbf{88.45}/86.91 & 66.67/\textbf{80.00} & 75.80/\textbf{82.67}
& \textbf{100.00}/96.67 & \textbf{100.00}/100.00 & \textbf{100.00}/98.18 
& - & - & - \\
SetActiveLimit 
& \textbf{96.08}/95.14 & 87.50/\textbf{93.75} & 91.37/\textbf{94.35} 
& \textbf{94.18}/93.00 & 81.82/\textbf{90.91} & 87.07/\textbf{91.76} 
& \textbf{100.00}/100.00 & \textbf{100.00}/100.00 & \textbf{100.00}/100.00 
& - & - & - \\
VerifyUserFace 
& \textbf{97.50}/68.50 & \textbf{86.96}/15.38 & \textbf{91.85}/24.99 
& \textbf{97.50}/68.50 & \textbf{86.93}/15.38 & \textbf{91.85}/24.99 
& - & - & - & - & - & - \\
SetAccountDailyQuota 
& 88.60/\textbf{92.21} & \textbf{82.86}/77.14 & \textbf{85.62}/83.92 
& 88.69/\textbf{95.47} & \textbf{81.74}/73.91 & \textbf{85.04}/83.18 
& \textbf{100.00}/100.00 & \textbf{100.00}/100.00 & \textbf{100.00}/100.00 
& \textbf{40.00}/30.00 & 40.00/\textbf{60.00} & \textbf{40.00}/40.00 \\
SetPayKey 
& \textbf{84.56}/71.12 & \textbf{80.00}/53.08 & \textbf{82.18}/60.73 
& \textbf{96.46}/90.97 & \textbf{91.11}/66.67 & \textbf{93.65}/76.88 
& \textbf{46.67}/19.33 & \textbf{48.00}/20.00 & \textbf{47.27}/19.64 
& 83.33/\textbf{87.50} & \textbf{66.67}/26.66 & \textbf{73.33}/38.00 \\
QueryPMAccount 
& 88.61/\textbf{92.82} & \textbf{80.69}/80.69 & 84.42/\textbf{86.23} 
& 87.68/\textbf{92.18} & \textbf{79.26}/79.26 & 83.20/\textbf{85.12} 
& \textbf{100.00}/100.00 & \textbf{100.00}/100.00 & \textbf{100.00}/100.00 
& - & - & - \\
OpenPSAccount
& \textbf{96.95}/94.54 & \textbf{93.06}/91.20 & \textbf{94.97}/92.84 
& \textbf{98.79}/94.65 & \textbf{97.58}/93.94 & \textbf{98.17}/94.29 
& \textbf{95.07}/93.99 & \textbf{88.72}/88.20 & \textbf{91.78}/91.00 
& \textbf{100.00}/100.00 & \textbf{100.00}/100.00 & \textbf{100.00}/100.00 \\

\midrule
Average
& \textbf{95.06}/89.84 & \textbf{89.97}/81.67 & \textbf{92.33}/83.41 
& \textbf{95.61}/91.39 & \textbf{89.56}/81.82 & \textbf{92.25}/84.09 
& \textbf{90.29}/85.00 & \textbf{89.45}/84.70 & \textbf{89.84}/84.80
& \textbf{74.44}/72.50 & \textbf{68.89}/62.22 & \textbf{71.11}/59.33 \\

\bottomrule
\end{tabular}
    }
\vspace{-1em}
\end{table*}

%% file: tables/comparison_of_MathPrompt.tex
\begin{table*}[htbp]
\centering
\vspace{1em}
\caption{Comparison of DDI task results with and without mathematical formalization prompting(MFP). The results are represented as data1/data2, where data1 and data2 refer to results with MFP and without MFP, respectively. The higher value is bolded.}
\label{tab:math_prompt_comparison}
\resizebox{\textwidth}{!}{
\begin{tabular}{@{}l|ccc|ccc|ccc|ccc@{}}
\toprule
\textbf{UseCase} 
& \multicolumn{3}{c|}{Overall} 
& \multicolumn{3}{c|}{API} 
& \multicolumn{3}{c|}{Condition} 
& \multicolumn{3}{c}{Action} \\
\cmidrule(lr){2-4} \cmidrule(lr){5-7} \cmidrule(lr){8-10} \cmidrule(lr){11-13}
& Precision & Recall & F1 
& Precision & Recall & F1 
& Precision & Recall & F1 
& Precision & Recall & F1 \\

\midrule
ClearFlag
& \textbf{100.00}/100.00 & \textbf{100.00}/100.00 & \textbf{100.00}/100.00 
& \textbf{100.00}/100.00 & \textbf{100.00}/100.00 & \textbf{100.00}/100.00 
& - & - & - & - & - & - \\
SetFlag
& \textbf{100.00}/100.00 & \textbf{100.00}/100.00 & \textbf{100.00}/100.00 
& \textbf{100.00}/100.00 & \textbf{100.00}/100.00 & \textbf{100.00}/100.00 
& - & - & - & - & - & - \\
QueryParentAccounts
& \textbf{100.00}/100.00 & \textbf{100.00}/100.00 & \textbf{100.00}/100.00 
& \textbf{100.00}/100.00 & \textbf{100.00}/100.00 & \textbf{100.00}/100.00 
& - & - & - & - & - & - \\
BindCard
& \textbf{100.00}/100.00 & \textbf{100.00}/100.00 & \textbf{100.00}/100.00 
& \textbf{100.00}/100.00 & \textbf{100.00}/100.00 & \textbf{100.00}/100.00 
& - & - & - & - & - & - \\
SetPassiveLimit
& \textbf{93.31}/91.52 & \textbf{78.57}/77.14 & \textbf{85.23}/83.70 
& \textbf{88.45}/85.23 & \textbf{66.67}/64.45 & \textbf{75.80}/73.33 
& \textbf{100.00}/100.00 & \textbf{100.00}/100.00 & \textbf{100.00}/100.00 
& - & - & - \\
SetActiveLimit
& \textbf{96.08}/93.13 & \textbf{87.50}/85.00 & \textbf{91.37}/88.81 
& \textbf{94.18}/89.50 & \textbf{81.82}/78.18 & \textbf{87.07}/83.30 
& \textbf{100.00}/100.00 & \textbf{100.00}/100.00 & \textbf{100.00}/100.00 
& - & - & - \\
VerifyUserFace
& \textbf{97.50}/95.87 & \textbf{86.96}/79.23 & \textbf{91.85}/83.46 
& \textbf{97.50}/95.87 & \textbf{86.93}/79.23 & \textbf{91.85}/83.46 
& - & - & - & - & - & - \\
SetAccountDailyQuota
& \textbf{88.60}/80.46 & \textbf{82.86}/70.00 & \textbf{85.62}/74.84 
& \textbf{88.69}/78.35 & \textbf{81.74}/67.83 & \textbf{85.04}/72.68 
& \textbf{100.00}/100.00 & \textbf{100.00}/100.00 & \textbf{100.00}/100.00 
& \textbf{40.00}/0.00 & \textbf{40.00}/0.00 & \textbf{40.00}/0.00 \\
SetPayKey
& 84.56/\textbf{85.97} & \textbf{80.00}/63.08 & \textbf{82.18}/72.51 
& \textbf{96.46}/93.24 & \textbf{91.11}/72.22 & \textbf{93.65}/80.90 
& 46.67/\textbf{60.00} & \textbf{48.00}/48.00 & 47.27/\textbf{52.78} 
& 83.33/\textbf{100.00} & \textbf{66.67}/33.33 & \textbf{73.33}/50.00 \\
QueryPMAccount
& 88.61/\textbf{94.86} & 80.69/\textbf{84.14} & 84.42/\textbf{89.00} 
& 87.68/\textbf{94.45} & 79.26/\textbf{82.96} & 83.20/\textbf{88.13} 
& \textbf{100.00}/100.00 & \textbf{100.00}/100.00 & \textbf{100.00}/100.00 
& - & - & - \\
OpenPSAccount
& \textbf{96.95}/96.88 & \textbf{93.06}/91.20 & \textbf{94.97}/93.94 
& \textbf{98.79}/95.08 & \textbf{97.58}/93.94 & \textbf{98.17}/94.50 
& 95.07/\textbf{98.31} & \textbf{88.72}/88.21 & 91.78/\textbf{92.90} 
& \textbf{100.00}/100.00 & \textbf{100.00}/100.00 & \textbf{100.00}/100.00 \\
\midrule
\textbf{Average}
& \textbf{95.06}/94.43 & \textbf{89.97}/86.34 & \textbf{92.33}/89.66 
& \textbf{95.61}/93.79 & \textbf{89.56}/85.35 & \textbf{92.25}/88.751 
& 90.29/\textbf{94.04} & 89.45/\textbf{90.89} & 89.84/\textbf{92.24}
& \textbf{74.44}/66.67 & \textbf{68.89}/44.44 & \textbf{71.11}/50.00 \\
\bottomrule
\end{tabular}
}
\vspace{1em}
\end{table*}

%% file: tables/comparison_of_context_purning.tex
\begin{figure*}[htbp]
    \centering
    \begin{subfigure}[b]{0.32\textwidth}
        \includegraphics[width=\textwidth]{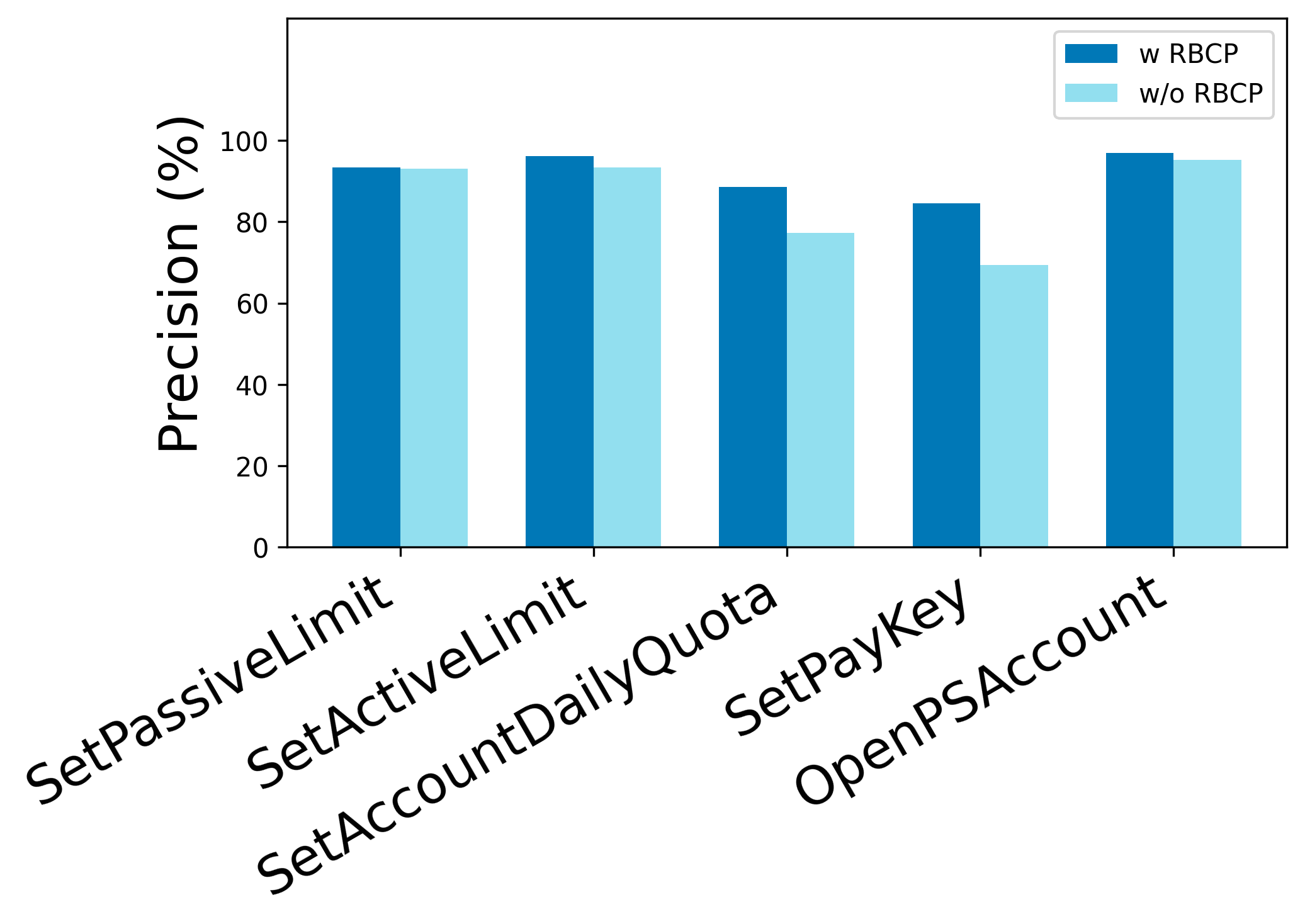}
        \caption{Precision}
    \end{subfigure}
    \hfill
    \begin{subfigure}[b]{0.32\textwidth}
        \includegraphics[width=\textwidth]{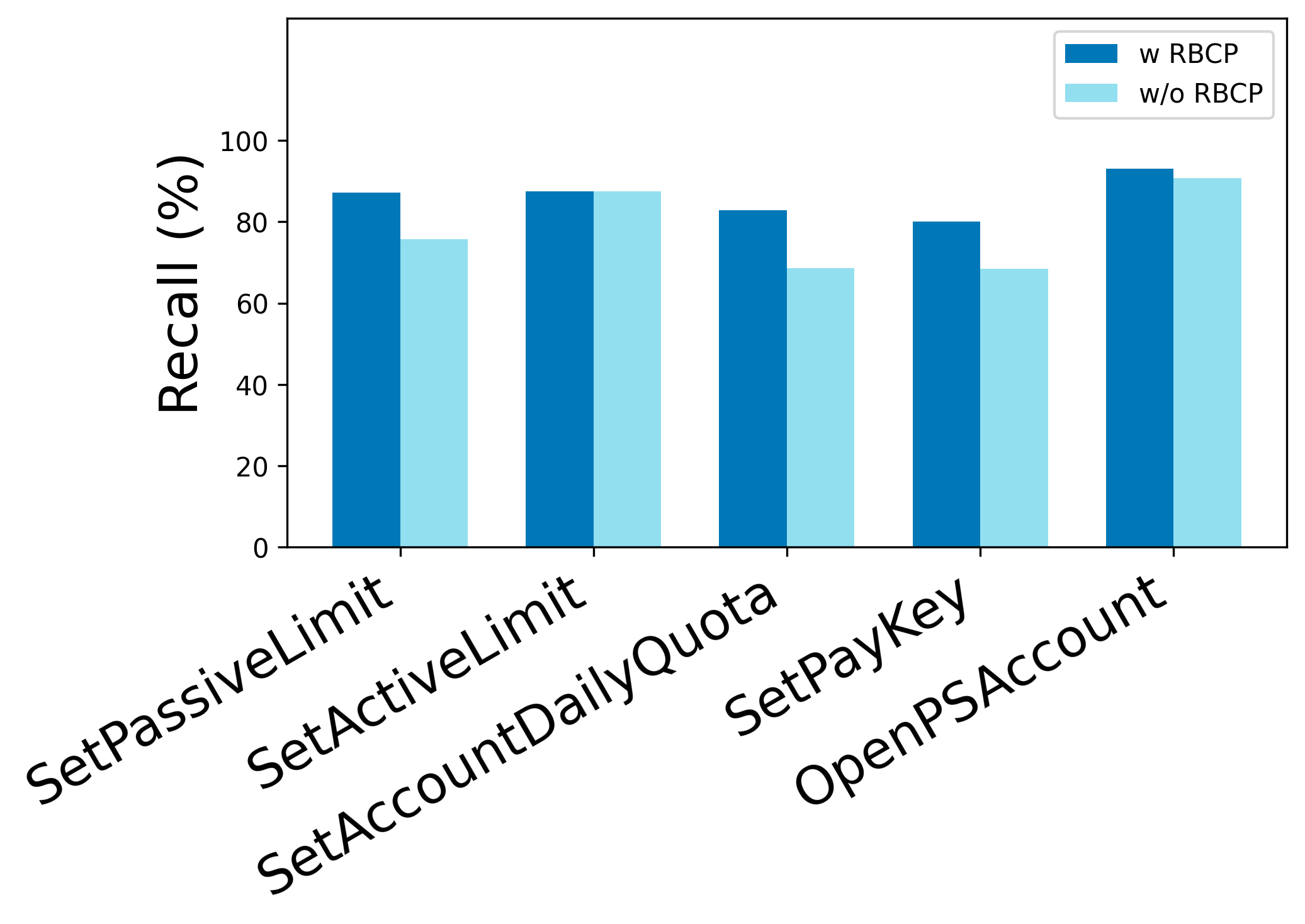}
        \caption{Recall}
    \end{subfigure}
    \hfill
    \begin{subfigure}[b]{0.32\textwidth}
        \includegraphics[width=\textwidth]{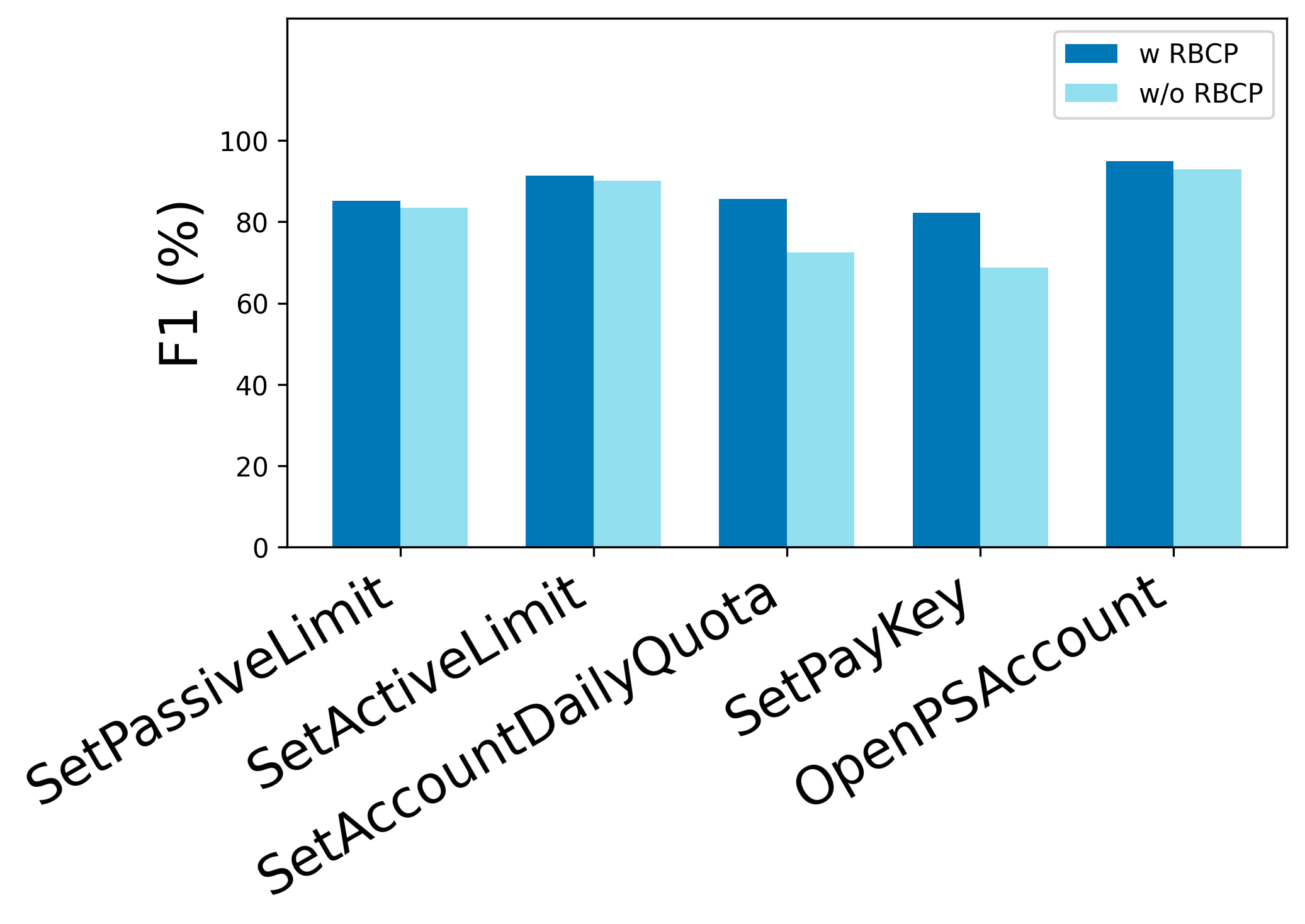}
        \caption{F1}
    \end{subfigure}
    \caption{Comparison of DDI task results with and without Reachability-Based Context Pruning(RBCP).}
    \label{fig:rbcp}
\end{figure*}

%% file: tables/codegen.tex
\begin{table*}[htbp]
\centering
\vspace{1em}
\caption{Comparison of compilation and unit test pass rates for each use case: w dependency, w/o dependency, and their difference (w dep - w/o dep). $\uparrow$ means increase, $\downarrow$ means decrease}
\label{tab:compile_unittest_comparison_diff_reordered}
\resizebox{0.75\textwidth}{!}{
\begin{tabular}{@{}l|lll|lll|lll@{}}
\toprule
\textbf{UseCase} 
& \multicolumn{3}{c|}{Compilation Pass Rate (\%)} 
& \multicolumn{3}{c|}{Full Unit Test Pass Rate (\%)} 
& \multicolumn{3}{c}{Unit Test Pass Rate (\%)} \\
\cmidrule(lr){2-4} \cmidrule(lr){5-7} \cmidrule(lr){8-10}
& w dep & w/o dep & Diff 
& w dep & w/o dep & Diff 
& w dep & w/o dep & Diff \\
\midrule
ClearFlag 
& 100 & 87 & $\uparrow$13
& 94 & 87 & $\uparrow$7
& 99 & 87 & $\uparrow$12 \\
SetFlag 
& 100 & 100 & \ \ 0
& 80 & 73 & $\uparrow$7
& 98 & 93 & $\uparrow$5 \\
QueryParentAccounts 
& 93 & 86 & $\uparrow$7
& 60 & 71 & $\downarrow$11
& 93 & 86 & $\uparrow$7 \\
BindCard 
& 100 & 100 & \ \ 0
& 100 & 100 & \ \ 0
& 100 & 100 & \ \ 0 \\
SetAccountDailyQuota 
& 73 & 69 & $\uparrow$4
& 27 & 0 & $\uparrow$27
& 70 & 60 & $\uparrow$10 \\
QueryPocketMoneyAccount 
& 100 & 71 & $\uparrow$29
& 0 & 0 & \ \ 0
& 97 & 61 & $\uparrow$36 \\
\midrule
Average
& 94.33 & 85.50 & $\uparrow$8.83
& 60.17 & 55.17 & $\uparrow$5.00
& 92.83 & 81.17 & $\uparrow$11.66 \\
\bottomrule
\end{tabular}
}
\vspace{1em}
\end{table*}

%% file: section/5_system_demonstration.tex
\section{SYSTEM DEMONSTRATION}

\begin{figure*}[htbp]
    \centering
    \begin{minipage}[t]{0.48\textwidth}
        \centering
        \includegraphics[width=\textwidth]{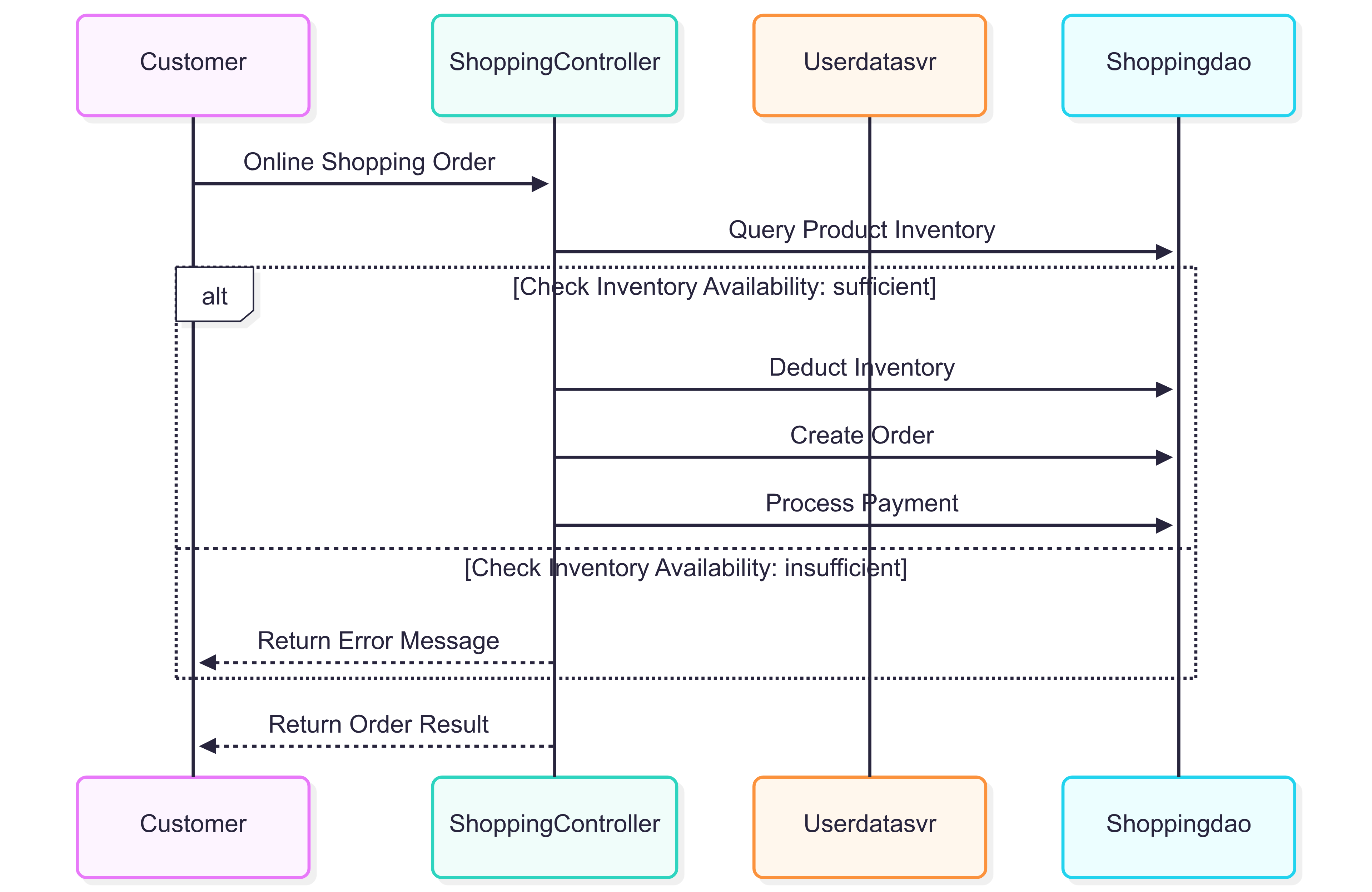}
        \caption{Example Sequence Diagram for Online Shopping}
        \label{fig:usecase}
    \end{minipage}%
    \hfill
    \begin{minipage}[t]{0.48\textwidth}
        \centering
        \includegraphics[width=\textwidth]{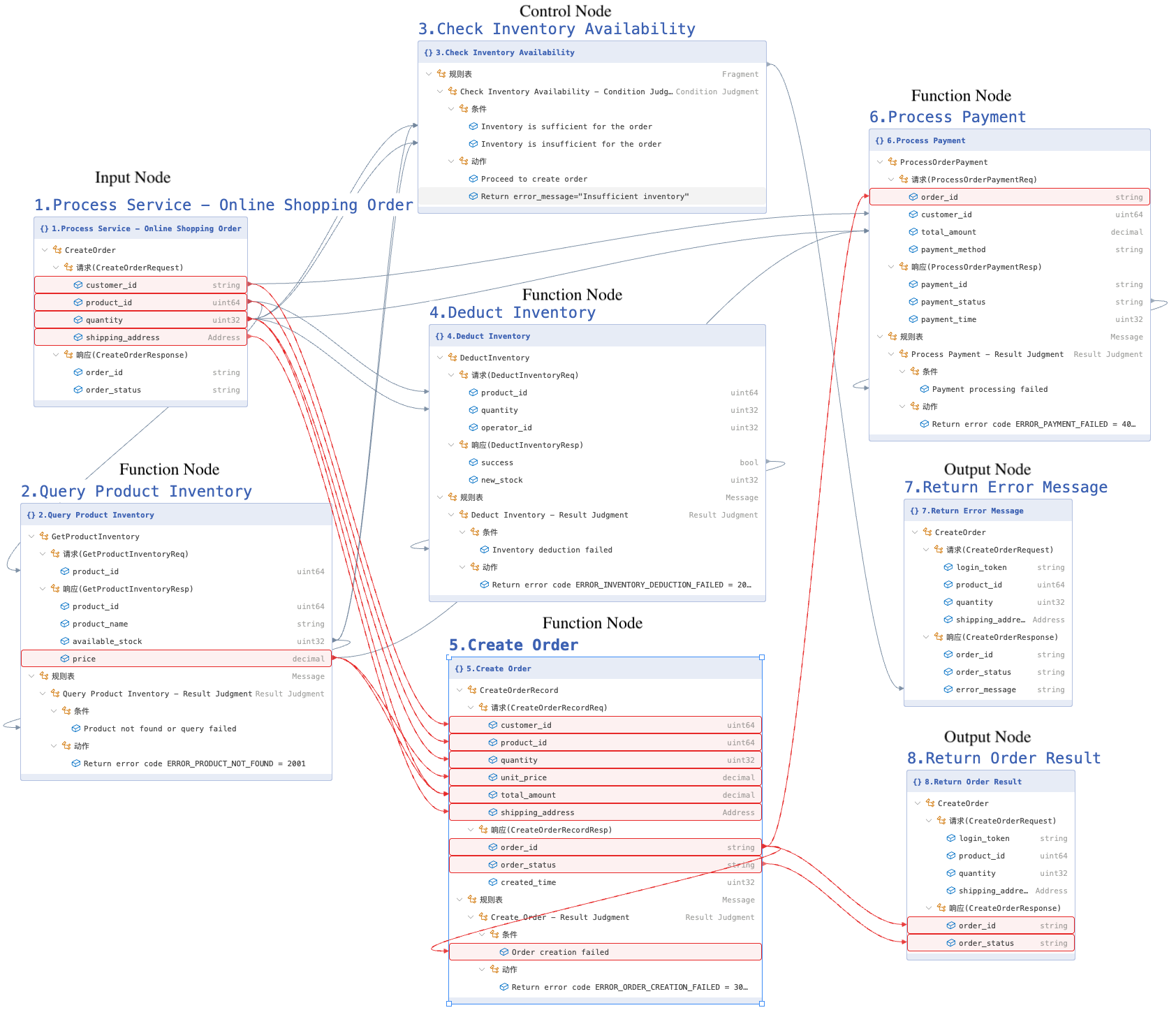}
        \caption{DDI Visualization Based on the Online Shopping Sequence Diagram}
        \label{fig:DDI}
    \end{minipage}
\end{figure*}

To facilitate early detection of design errors and enhance modeling efficiency, we encapsulate the proposed DDI framework as an AI-assisted tool and integrate it into the sequence diagram design system. 

Figure \ref{fig:usecase} presents a sequence diagram for an online shopping scenario. The main workflow includes querying product inventory, creating an order, and processing payment. Exception handling for insufficient inventory is also depicted, introducing alternative business flows. Due to its complexity and multiple branches, this sequence diagram effectively demonstrates the capabilities of our Data Dependency Inference (DDI) task.

Figure \ref{fig:DDI} visualizes the inferred data dependency relationships. During sequence diagram construction, engineer can invoke data dependency inference, after which the system visualizes all inferred data dependency edges. This visual interface allows engineer to intuitively identify potential issues, such as ambiguities in decision table definitions or omissions in data flow, thereby achieving a tight integration of dependency inference and design validation.

The dependency inference system supports flexible analysis scopes. Engineer can perform global dependency inference across the entire sequence diagram to comprehensively assess the rationality of data dependencies, or conduct local analysis on specific nodes (e.g., function nodes, control nodes, or output nodes). For example, after completing the sequence diagram, a global inference can be executed to promptly detect errors. During modeling, engineer may also infer dependencies for the currently edited node, enabling real-time validation and timely correction based on system-generated warnings or errors. Additionally, when issues arise at a particular step, targeted inference can be performed on the relevant node, allowing engineer to leverage system feedback for precise localization and resolution.

During dependency inference, the system automatically analyzes the provenance of each data item, verifies whether predecessor nodes provide the required data, and checks for necessary data type conversions or other processing. For instance, if a consumer requires a \texttt{user\_id} of type \texttt{uint64}, but the provider supplies a \texttt{user\_id} of type \texttt{uint32}, the system issues a type compatibility warning, prompting the engineer to consider type conversion. If no data source is identified among predecessor nodes, the system generates an error message indicating a missing data source and advises the engineer to review prerequisite operations or nodes.


%% file: section/6_related_work.tex
\section{RELATED WORK}

\subsection{UML Modeling and Analysis}
The Unified Modeling Language (UML) is a standardized, general-purpose modeling language designed to represent the interactions among collaborating objects in software systems~\cite{booch1996unified, ozkaya2020survey, hammond2006tahuti}. Over the past two decades, a variety of approaches leveraging UML for modeling and analysis tasks have been developed.

Early work focused on reverse engineering, which can be categorized into static and dynamic approaches\cite{briand2006toward, abidi2015new}. 
A seminal contribution in static analysis is the study by Rountev et al~\cite{rountev2005static}, which presents an algorithm for mapping reducible, exception-free control-flow graphs to UML interaction fragments. 
Conversely, dynamic approaches~\cite{delamare2006reverse, sarkar2013reverse, richner2002using, oechsle2002javavis} primarily focus on analyzing application performance through execution traces to generate sequence diagrams. These methods often utilize runtime data to reconstruct behavioral models that reflect actual system operations. 

Reverse engineering reconstructs UML from existing code. Subsequent innovations have introduced more sophisticated frameworks. Notably, aToucan~\cite{aToucan} proposes a rule-driven end-to-end system capable of automatically generating UML analysis models from use case descriptions written in constrained natural language. 
Similarly, Jahan et al.~\cite{jahan2021generating} propose an automated method to generate UML sequence diagrams from textual use cases.

Recent studies have further explored the application of LLMs. Ferrari et al.~\cite{FerrariAA24} systematically evaluate the capabilities of GPT-3.5 and GPT-4 in generating UML sequence diagrams directly from natural language requirements. 
However, it also identifies a critical gap in addressing Data Dependency Inference(DDI), a key issue that our paper addresses.

\subsection{From Design to Code: Challenges and Industrial Reality}
The automated translation of software designs into executable code remains a persistent challenge in software engineering, particularly when scaling to complex enterprise systems with intricate data dependencies.

Early model-driven approaches like UJECTOR \cite{usman2008ujector} established foundational workflows for structural code generation from UML class diagrams and basic behavioral patterns from sequence/activity diagrams. Subsequent frameworks \cite{kluisritrakul2016generation} improved standardization through XMI parsing and modular rule mapping. 

The emergence of LLMs has introduced new paradigms. Sadik et al\cite{DBLP:conf/modelsward/SadikBO24} leverages GPT-4 to bridge UML/OCL specifications with code generation, demonstrating enhanced adaptability compared to traditional template-based approaches. Recent multimodal approaches \cite{bates2025unified} attempt to address visual design artifacts through image-based UML processing.

%% file: section/7_conclusion.tex
\section{CONCLUSION}
We propose a systematic framework for Data Dependency Inference from industrial UML sequence diagrams, combining enhanced sequence diagram, mathematical formalization prompting, and reachability-based context pruning. Our framework enables LLMs to accurately infer data dependencies, significantly improving both the quality of code generation and the reliability of design verification. Experiments on real-world microservice use cases show that our method achieves high precision and recall across multiple dependency types, with an average
recall of 89.97\%. This framework not only streamlines automated code synthesis but also serves as an effective tool for sequence diagram validation in complex software engineering practice.

%% file: section/8_ack.tex
\section*{Acknowledgments}
We appreciate the assistance from the WeChat pay team for their valuable contributions. This research is supported by the National Natural Science Foundation of China under project (No. 62472126, 62276075, 62192731), Natural Science Foundation of Guangdong Province (Project No. 2023A1515011959), Shenzhen-Hong Kong Jointly Funded Project (Category A, No. SGDX20230116 091246007) and the Tencent WeChat Rhino-Bird Focused Research Program.